\documentclass[letterpaper, 10 pt, conference]{ieeeconf}  
\IEEEoverridecommandlockouts                             
\usepackage{graphics}  
\usepackage{times}     
\usepackage{amsmath}   
\usepackage{amssymb}   
\usepackage{graphicx}
\usepackage{algorithm}
\usepackage[noend]{algpseudocode}
\usepackage{siunitx}
\usepackage{multirow}
\usepackage{booktabs}
\usepackage{xspace}
\usepackage{xcolor}
\usepackage[normalem]{ulem}
\usepackage{xfrac}

\usepackage[font=small]{caption}

\def\secref#1{Sec.~\ref{#1}}
\def\figref#1{Fig.~\ref{#1}}
\def\tabref#1{Tab.~\ref{#1}}
\def\eqref#1{Eq.~(\ref{#1})}

\usepackage{color}

\def\etal{\textit{et al.\,}}

\newcommand{\cropvsweed}{crop-weed\xspace}
\newcommand{\Cropvsweed}{Crop-weed\xspace}

\newcommand{\sequentialModule}{sequential module\xspace}
\newcommand{\visualEncoder}{visual encoder\xspace}
\newcommand{\visualDecoder}{visual decoder\xspace}
\newcommand{\visualDecoderOutput}{visual features\xspace}
\newcommand{\visualCode}{visual code\xspace}
\newcommand{\sequenceCode}{sequence code\xspace}
\newcommand{\visualCodes}{visual codes\xspace}
\newcommand{\fusion}{spatio-temporal fusion\xspace}
\newcommand{\sequenceDecoder}{spatio-temporal decoder\xspace}
\newcommand{\sequenceDecoderOutput}{sequence features\xspace}

\usepackage{array}
\newcolumntype{L}[1]{>{\raggedright\let\newline\\\arraybackslash\hspace{0pt}}m{#1}}
\newcolumntype{C}[1]{>{\centering\let\newline\\\arraybackslash\hspace{0pt}}m{#1}}
\newcolumntype{R}[1]{>{\raggedleft\let\newline\\\arraybackslash\hspace{0pt}}m{#1}}

\usepackage{fancyhdr}
\title{\LARGE \bf  Fully Convolutional Networks with Sequential Information\\ for Robust Crop and Weed Detection in Precision Farming}

\author{Philipp Lottes \and Jens Behley  \and Andres Milioto \and Cyrill Stachniss%
  \thanks{All authors are with the University of Bonn, 
  Institute of Geodesy and Geoinformation, Bonn, Germany. This work has partly 
  been supported by the EC under the grant number H2020-ICT-644227-Flourish.}
}

\begin{document}
\maketitle
\thispagestyle{fancy}
\pagestyle{fancy}

\begin{abstract}
Reducing the use of agrochemicals is an important component towards  
sustainable agriculture. Robots that can perform targeted weed
control offer the potential to contribute to this goal, for example, through
specialized weeding actions such as selective spraying or mechanical weed removal. 
A~prerequisite of such systems is a reliable and robust plant classification
system that is able to distinguish crop and weed in the field. A major
challenge in this context is the fact that different fields show a large variability.
Thus, classification systems have to robustly cope with substantial
environmental changes with respect to weed pressure and weed types, growth
stages of the crop, visual appearance, and soil conditions.

In this paper, we propose a novel \cropvsweed classification system that
relies on a fully convolutional network with an encoder-decoder structure and
incorporates spatial information by considering image sequences.
Exploiting the crop arrangement information that is observable from the image
sequences enables our system to robustly estimate a pixel-wise labeling of the
images into crop and weed, i.e.\,, a semantic segmentation. We provide a thorough experimental
evaluation, which shows that our system generalizes well to previously unseen
fields under varying environmental conditions---a key capability to
actually use such systems in precision framing. We provide comparisons to
other state-of-the-art approaches and show that our system substantially
improves the accuracy  of \cropvsweed classification without requiring a
retraining of the model.

\end{abstract}

\section{Introduction}
\label{sec:intro}

\begin{figure}
 \centering
 \includegraphics[width=0.92\linewidth]{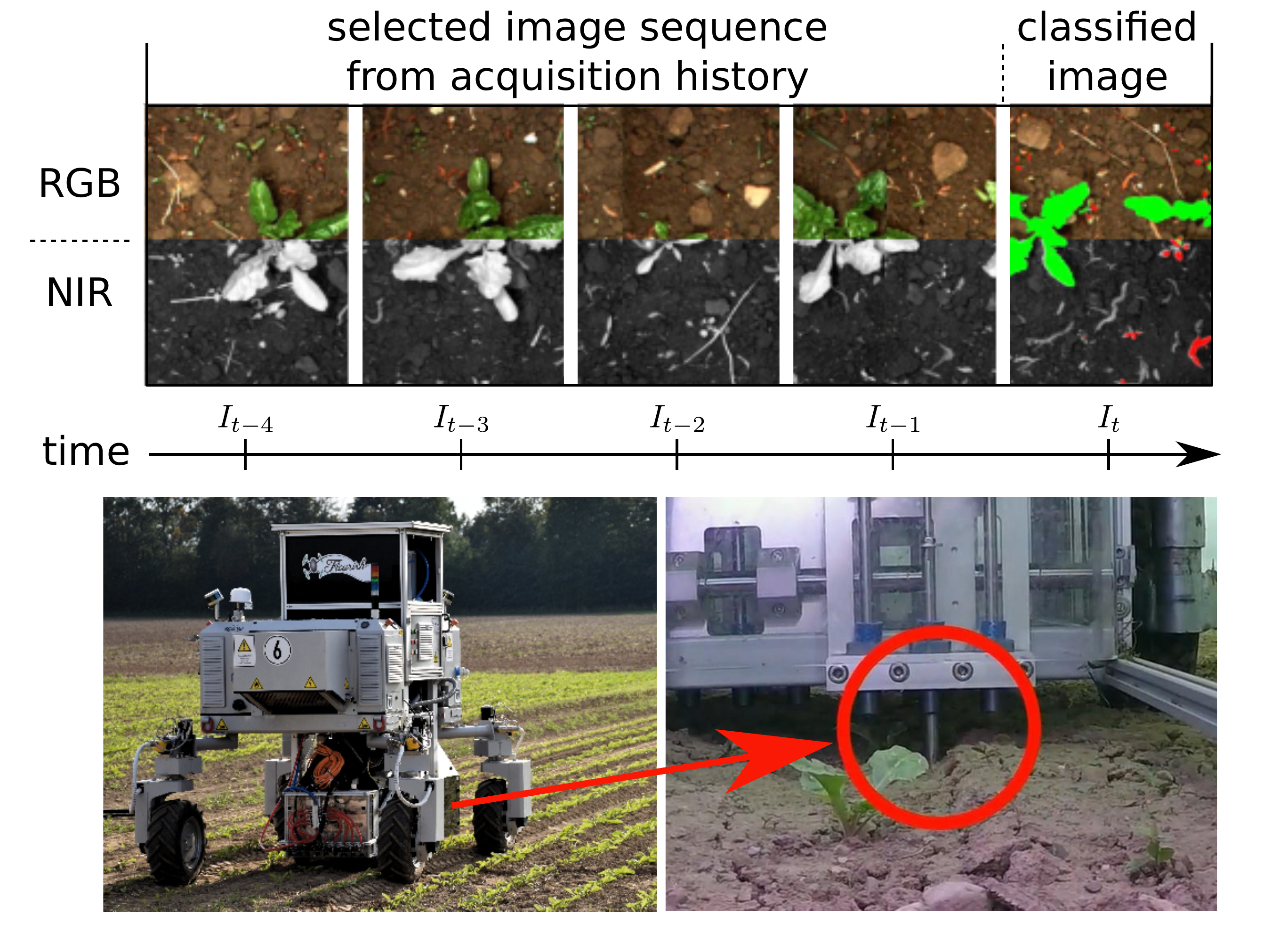}
\caption{The image depicts the BoniRob platform perceiving RGB+NIR
images along the crop row as well as \cropvsweed classification results by
exploiting the repetitive plant arrangement information. Our approach analyzes
image sequences from the acquisition history and provides a pixel-wise image
labeling of the field under the robot. The lower right image shows a mechanical actuator stamping
the detected weeds into the ground.}
\label{fig:motivation}
\end{figure}

A sustainable agriculture is one of the seventeen sustainable development
goals of the United Nations. A key objective is to reduce the reliance on
agrochemicals such as herbicides or pesticides due to its side-effects on the
environment, biodiversity, and partially human health.  Autonomous precision
farming robots have the potential to address this challenge by performing
targeted and plant-specific interventions in the field. These
interventions can be selective spraying or mechanical weed stamping, see
\figref{fig:motivation}. A crucial prerequisite for selective treatments
through robots is a plant classification system that enables robots to
identify individual crop and weed plants in the field. This is the basis for
any following targeted treatment action
\cite{mueter2013aec,Langsenkamp2014cigr}.

Several vision-based learning methods have been proposed in the context of
crop and weed detection~\cite{haug2014wacv,lottes2017icra,milioto2018icra,
mortensen2016cigr,nieuwenhuizen2009phd,potena2016ias}. Typically, such methods
are based on supervised learning methods and report high classification
performances in the order of $80$-$95$\% in terms of accuracy. However, we see a
certain lack in the evaluation of several methods with respect to the
generalization  capabilities to unseen situations, i.e., the classification
performance under substantial changes in the plant appearance and soil
conditions between training and testing phase. This generalization
capabilities to new fields, however, is essential for actually deploying precision
farming robots with selective intervention capabilities.
Also Slaughter \etal~\cite{slaughter2008cea} conclude in their review about robotic 
weed control systems that the missing generalization capabilities of the \cropvsweed 
classification technology is a
major problem with respect to the deployment and commercialization of such
systems. Precision farming robots need to operate in different fields on a
regular basis. Thus, it is crucial for the classification system to provide
robust results under changing conditions such as differing weed density
and weed types as well as appearance, growth stages of the  plants, and
finally soil conditions and types. Purely visual \cropvsweed classification
systems often degrade due to such environmental changes as these 
correlate with the underlying data distribution of the images.

In this work, we aim at overcoming the performance loss in \cropvsweed
detection due to changes in the underlying distribution of the image data. To
achieve that, we exploit geometric patterns that result from the fact that
several  crops are sowed in rows. Within a field of row crops (such as sugar beets or
corn), the plants share a similar lattice distance along the row,
whereas weeds appear more randomly. In contrast to the visual
cues, this  geometric signal is much less affected by changes in the visual
appearance. Thus, we propose an approach to exploit this information as an
additional signal by analyzing image sequences that cover a local strip of the
field surface in order to improve the classification performance.
\figref{fig:motivation} depicts the employed farming robot during operation in
the field and an example of an image sequence consisting of
4-channel images, i.e., conventional red, green, and blue~(RGB) channels as
well as an additional near infra-red~(NIR) channel.

The main contribution of this work is a novel, vision-based \cropvsweed
classification approach that operates on image sequences. Our
approach performs pixel-wise semantic segmentation of images into soil, crop,
and weed.  Our proposed method is based on a fully convolutional network
(FCN) with an encoder-decoder structure and incorporates 
information about the plant arrangement by a novel extension. This extension,
called sequential module, enables the usage of image sequences to implicitly
encode the local geometry. This combination leads to better generalization
performance even if the visual appearance or the growth stage of the plants
changes between training and test time. Our system is trained end-to-end
and relies neither on pre-segmentation of the vegetation nor on any kind of
handcrafted features.

We make the following two key claims about our approach. First, it generalizes
well to data acquired from unseen fields and robustly classifies crops at
different growth stages without the need for retraining (we achieve an average
recall of more than $94$\% for crops and over $91$\% for weeds even in
challenging settings). Second, our approach is able to extract features about
the spatial arrangement of the plantation from image sequences and can exploit
this information to detect crops and weeds solely based on geometric cues.
These two claims are experimentally validated on real-world datasets, which
are partially publicly available~\cite{chebrolu2017ijrr}. We compare our
approach to other state-of-the-art approaches and to a non-sequential FCN
model. Through simulations, we furthermore demonstrate the approach's ability
to learn plant arrangement information. We plan to publish our code.

\section{Related Work}
\label{sec:related}

While there has recently been significant progress towards robust vision-based
\cropvsweed classification, many systems rely on handcrafted features
\cite{haug2014wacv, lottes2016jfr,lottes2017icra}. However, the
advent of convolution neural networks (CNN) also spurred increasing interest
in end-to-end \cropvsweed classification systems \cite{cicco2017iros, mccool2017ral, milioto2017uavg, milioto2018icra, mortensen2016cigr,potena2016ias} to overcome the inflexibility and limitations of handcrafted
vision pipelines.

In this context, CNNs are often applied in pixel-wise fashion operating on
image patches provided by a sliding window approach. Using this principle,
Potena \etal~\cite{potena2016ias} use a cascade of CNNs for \cropvsweed
classification, where the first CNN detects vegetation 
and then only the vegetation pixels are classified by a deeper \cropvsweed
CNN. McCool \etal~\cite{mccool2017ral} fine-tune a very deep
CNN \cite{szegedy2016cvpr} and attain practical processing
times by compression of the fine-tuned network using a mixture of small, but fast
networks, without sacrificing too much classification accuracy.

Fully convolutional networks (FCN) \cite{long2015cvpr-fcnf} directly estimate
a pixel-wise segmentation of the complete image and can therefore use
information from the whole image. The encoder-decoder architecture  of 
SegNet \cite{badrinarayanan2017pami} is nowadays a common building
block of semantic segmentation approaches
\cite{paszke2016arxiv,ronneberger2015micc}. Sa \etal
\cite{sa2018ral} propose to use the SegNet architecture for multi-spectral
\cropvsweed classification from a micro aerial vehicle.
Milioto \etal~\cite{milioto2018icra} use a similar architecture, but
combine RGB images with background knowledge encoded in additional
input channels. 

All systems share the need for tedious re-labeling effort to
generalize to a different field. To alleviate this burden, Wendel and
Underwood \cite{wendel2016icra} enable self-supervised learning by exploiting
plant arrangement in rows. Recently, we \cite{lottes2017iros} proposed to
exploit the plant arrangement to re-learn a random forest with minimal 
labeling effort in a semi-supervised way. Hall \etal~\cite{hall2017iros}
minimize the labeling effort through an unsupervised weed scouting process
\cite{hall2017icra} that clusters visually similar images and allows to
assign labels only to a small number of candidates to label the whole acquired
data. Cicco \etal \cite{cicco2017iros} showed that a SegNet-based
\cropvsweed segmentation can be learned using only synthetic images.

In contrast to the aforementioned prior work, we combine a FCN with sequential
information to exploit the repetitive structure of the plant arrangement
leading to better generalization performance without re-labeling effort. To
the best of our knowledge, we are therefore the first to propose an end-to-end
learned semantic segmentation approach exploiting the spatial arrangement of
row plants.

\section{Sequential \Cropvsweed Classification System}
\label{sec:main}

\begin{figure*}[t]
 \centering
 \includegraphics[width=0.92\linewidth]{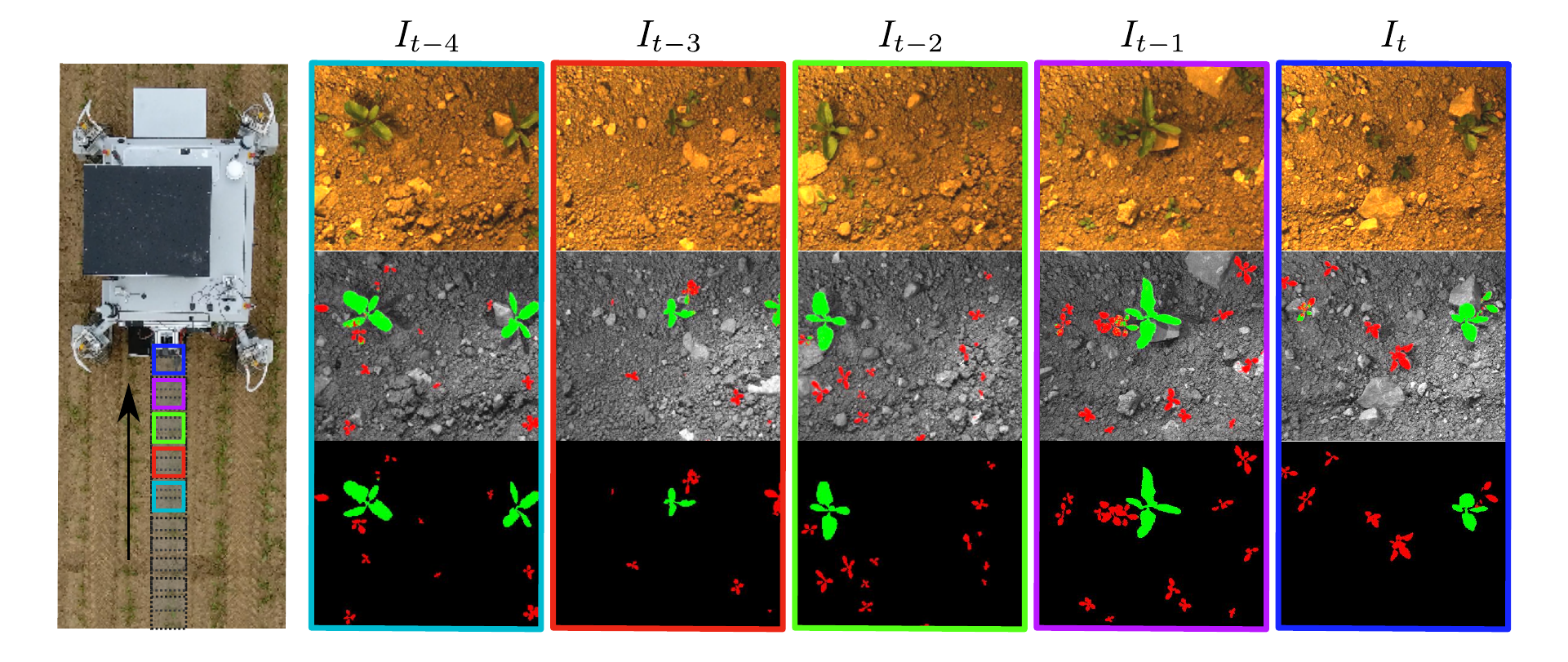}
\vspace{-2mm}
 \caption{Left: BoniRob acquiring images while driving along the crop row.
 Our approach exploits an image sequence by selecting those image from the
 history that do not overlap in object space. Right: Exemplary prediction of
 crop and weed for an image sequence captured on a field near Stuttgart where
 the classification model was solely trained on data acquired in Bonn; top
 row: RGB images; middle  row:  predicted label mask projected on the NIR
 image (crop in green, weed in red, background transparent); bottom row:
 ground truth.
 }
  \label{fig:sequence}
  \vspace{-0.15cm}
\end{figure*}

The primary objective of our approach is to enable robots to \emph{robustly}
identify crops and weeds during operation in the field, even when the visual
appearance of the plants and soil has changed. We propose a sequential 
FCN-based classification model for pixel-wise semantic segmentation of the classes
(i) background, i.e. mostly soil, (ii) crops (sugar beet), and (iii) weeds.
The key idea is to exploit information about the plant arrangement, which is
given by the planting process. We let the classification model learn this
information from image sequences representing a part of the crop row and fuse
the arrangement information with visual features. With this novel combination,
we can improve the performance and generalization capabilities of our
classification system.

\figref{fig:sequence} depicts the $S=5$ selected RGB+NIR images for
building the sequence $\mathcal{I} = \{I_t, \ldots, I_{t-4}\}$ as input to our
pipeline as well as exemplary predictions and their corresponding ground truth
label masks. To exploit as much spatial information as possible with a small
number of images, we select those images along the traversed trajectory that
do not overlap in object space. For the image selection procedure, we use the
odometry information and the known calibration parameters of the camera. The
rightmost image $I_t$ refers to the current image, whereas $I_{t-1},
\ldots, I_{t-4}$ are selected non-overlapping images from the history of acquired images. 
The output of the proposed network is a label mask
representing a probability distribution over the class labels for each pixel
position.

\subsection{General Architectural Concept}
\label{sec:archconcept}

\begin{figure*}[t]
  \centering

  \includegraphics[width=0.97\linewidth]{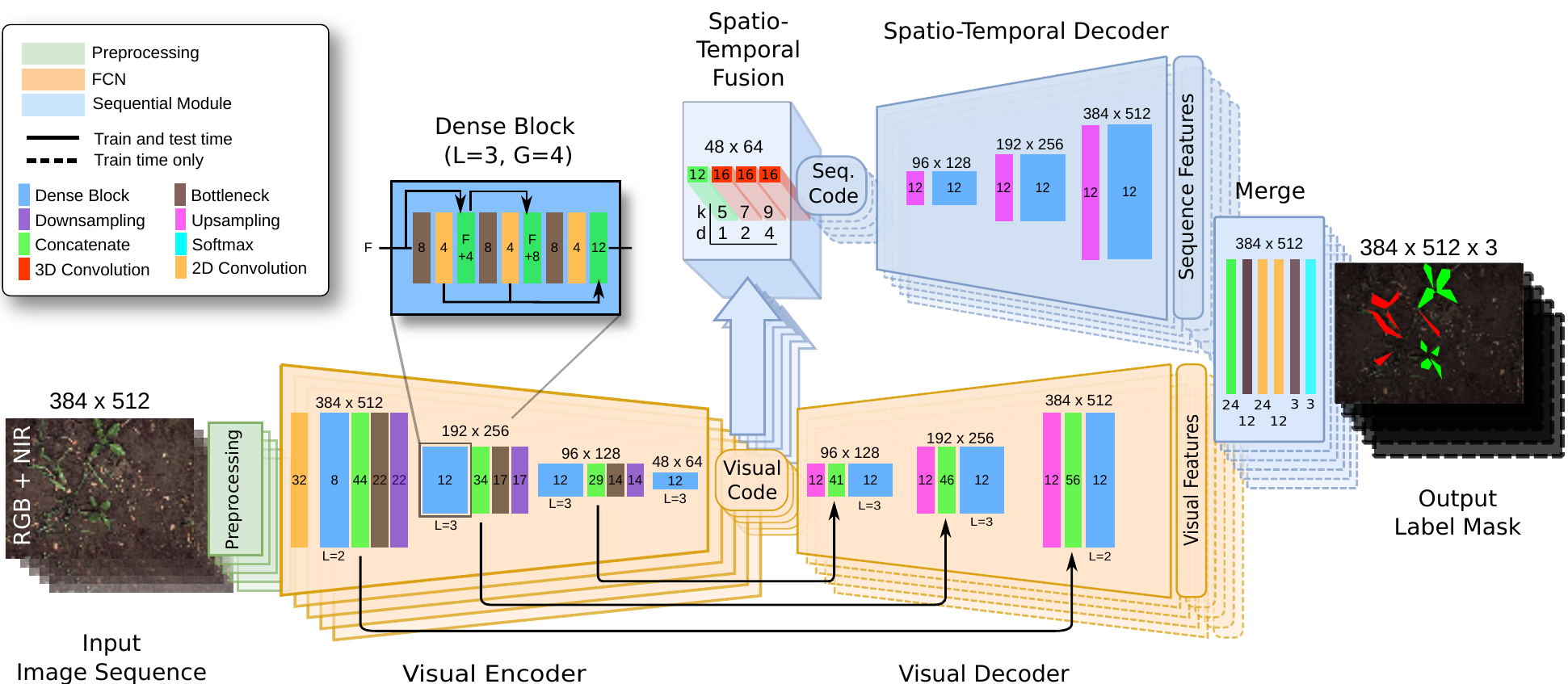}
  
  \caption{Architecture of our approach as described in \secref{sec:archconcept}. 
    Given a sequence of $S$ images, we first compute the \visualCode through the \visualEncoder.
    The \visualCode is then again upsampled with the \visualDecoder resulting in the \visualDecoderOutput.
    The \fusion uses $S$ \visualCodes to encode the sequential context from multiple images resulting in the \sequenceCode.
    Using the \sequenceDecoder, we upsample the \sequenceCode again resulting in the \sequenceDecoderOutput.
    Finally, the visual and \sequenceDecoderOutput are merged resulting in a pixel-wise label mask.
    In the figure, we included the size of the feature maps above the layers and the number of feature maps produced by each layer inside the layer.
    In the spatio-temporal fusion, we included the parameters $k$ and $d$ of the dilated convolution.} 
\label{fig:archconcept}
\vspace{-0.15cm}
\end{figure*}

\figref{fig:archconcept} depicts the conceptual graph and the information flow
from the input to the output for a sequence $\mathcal{I}$ of length $S = 5$.
We divide the model into three main blocks: (i)~the~preprocessing block
(green), (ii)~the encoder-decoder FCN (orange), and (iii)~the sequential module
(blue). We extend the encoder-decoder FCN with our proposed
\sequentialModule. Thus, we transform the FCN model for single images into a
sequence-to-sequence model.

The \visualEncoder of the FCN shares its weights along the time axis such that
we reuse it as a task-specific feature encoder for each image separately.
This leads to the computation of $S$~\visualCodes being a compressed, but highly
informative representation of the input images. We route the \visualCode along
two different paths within the architecture. First, each \visualCode is passed
to the decoder of the encoder-decoder FCN sharing also its weights along the
time axis. This leads to $S$ decoded \visualDecoderOutput volumes. Thus, the
encoder-decoder FCN is applied to each images separately. Second, all
\visualCode volumes of sequence $\mathcal{I}$ are passed to the \sequentialModule. The
\sequentialModule processes the $S$ \visualCodes jointly as a sequence by
using 3D convolutions and outputs a \sequenceCode, which contains information
about the sequential content. The \sequenceCode is then passed through a
\sequenceDecoder to upsample it to the same image resolution as the
\visualDecoderOutput, the \sequenceDecoderOutput. The resulting visual feature maps and the
sequential feature maps are then merged to obtain the desired label mask
output.

\subsection{Preprocessing}
\label{sec:preprocessing}

Preprocessing the input can help to improve the generalization capabilities of
a classification system by aligning the training and test data distribution.
The main goal of our preprocessing is to minimize the diversity of the
inputs as much as possible in a data-driven way. We perform the preprocessing
independently for each image and moreover separately on all channels, i.e.
red, green, blue, and near infra-red. For each channel, we (i)~remove noise by
performing a blurring operation using a $[5\times5]$ Gaussian kernel given by
the standard normal distribution, i.e., $\mu = 0$ and $\sigma^2 = 1$,
(ii)~standardize the data by the mean and standard deviation, and 
(iii)~normalize and zero-center the values to the interval $[-1,1]$ as it is
common practice for training FCNs. \figref{fig:dataset} illustrates 
the effect of our preprocessing for exemplary images 
captured with different sensor setups.

\subsection{Encoder-Decoder FCN}
\label{sec:baseFCN}

It has been shown that FCNs for semantic segmentation tasks achieve high
performance for a large number of different
applications~\cite{badrinarayanan2017pami,paszke2016arxiv,ronneberger2015micc,
huang2017cvpr-dccn}. Commonly, FCN architectures follow the so-called
``hourglass'' structure referring to a downsampling of the resolution in the
encoder followed by a complementary upsampling in the decoder to regain the full
resolution of the input image for the pixel-wise segmentation. We design our
network architecture keeping in mind that the classification needs to run in
near real-time such that an actuator can directly act upon the incoming
information. In addition to that we use a comparably small number of learnable
parameters to obtain a model capacity which is sufficient for the three class
prediction problem. Thus, we design a lightweight network for our task.

As a basic building block in our encoder-decoder FCN, we follow the so-called
Fully Convolutional DenseNet (FC-DenseNet) \cite{jegou2017arxiv}, which
combines the recently proposed densely connected CNNs organized as dense
blocks \cite{huang2017cvpr-dccn} with fully convolutional networks (FCN)
\cite{long2015cvpr-fcnf}. The key idea is a dense connectivity pattern which
iteratively concatenates all computed feature maps of subsequent convolutional
layers in a feed forward fashion. These ``dense'' connections encourage deeper
layers to reuse features produced by earlier layers and
additionally supports the gradient flow in the backward pass.

As commonly used in practice, we define our 2D convolutional layer as a
composition of the following components: (1) 2D convolution, (2) rectified
linear unit (ReLU) as non-linear activation, (3) batch normalization
\cite{szegedy2016cvpr} and (4) dropout \cite{srivastava2014jmlr}. We
repeatedly apply bottleneck layers to the feature volumes and thus keep the
number of feature maps small while achieving a deep architecture. Our
bottleneck is a 2D convolutional layer with a $[1\times1]$ kernel.

A dense block is given by a stack of $L$~subsequent 2D convolutional layers
operating on feature maps with the same spatial resolution.
\figref{fig:archconcept} depicts the information flow in a dense block. The
input of the $l^\text{th}$ 2D convolutional layer is given by a concatenation
of all feature maps produced by the previous layers, whereas the output
feature volume is given by the concatenation of the \emph{newly} computed
feature maps within the dense block. Here, all the concatenations are
performed along the feature axis. The number of the produced feature maps is
called the growth rate $G$ of a dense block \cite{huang2017cvpr-dccn}. We
consequently use $2G$ kernels for the bottleneck layers within a dense block
to reduce the computational cost in the subsequent 2D convolutional layers.

\figref{fig:archconcept} illustrates the information flow through the FCN. The
first layer in the encoder is a 2D convolutional layer augmenting the
4-channel images using $32$ $[5\times5]$ kernels. The following operations in
the encoder are given by a recurring composition of dense blocks, bottleneck
layers and downsampling operations, where we concatenate the input of a dense
block with its output feature maps. We perform the downsampling by strided
convolutions employing 2D convolutional layers with an $[5\times5]$ kernel and
a stride of~$2$. All bottleneck layers between dense blocks compress the
feature volumes by a learnable halving along the feature axis. In the decoder,
we revert the downsampling by a strided transposed convolution
\cite{dumoulin2016arxiv} with a $[2\times2]$ kernel and a stride of $2$. To
facilitate the recovery of spatial information, we concatenate feature maps
produced by the dense blocks in the encoder with the corresponding feature
maps produced by the learnable upsampling and feed them into a bottleneck
layers followed by a dense blocks. Contrary to the encoder, we reduce the
expansion of the number of feature maps within the decoder by omitting the
concatenation of a dense blocks input with its respective output.

\subsection{Sequential Module}
\label{sec:sm}

To learn information about the crop and weed arrangement, the network 
needs to take the whole sequence corresponding to a crop row
into account. The \sequentialModule represents our key architectural design
contribution to enable sequential data processing. It can be seen as an
additional parallel pathway for the information flow and consists of three
subsequent parts, i.e. the (i)~\fusion, the (ii)~\sequenceDecoder and the
(iii)~merge layer.

The \fusion is the core part of the sequential processing. First, we create a
sequential feature volume by concatenating all visual code volumes of the
sequence along an additional time dimension. Second, we compute a 
spatio-temporal feature volume, the \sequenceCode, as we process the built
sequential feature volume by a stack of three 3D convolutional layers. We
define the 3D convolutional layer analogous to the 2D convolutional layer,
i.e. a composition of convolution, ReLu, batch normalization and dropout. In
each 3D convolutional layer, we use $16$ 3D kernels with a size of
$[5\times5\times S]$ to allow the network to learn weight updates under
consideration of the whole input sequence. We apply the batch normalization to
all feature maps jointly regardless of their position in the sequence.

As a further architectural design choice, we increase the receptive field for
subsequent applied 3D convolutional layers in their spatial domain. Thus, the
network can potentially exploit more context information to extract the
geometric arrangement pattern of the plants. To achieve this, we increased the
kernel size $k$ and the dilation rate $d$ of the 3D kernels for subsequent 3D
convolutional layers. More concretely, we increase $k$ and $d$ only for the
spatial domain of the convolutional operation, i.e. $[k\times k\times S]$ with
$k=\{5,7,9\}$ and $[d\times d\times 1]$ with $d=\{1,2,4\}$.
This leads to a larger receptive field of the \fusion allowing the model to
consider the whole encoded content of all images along the sequence. In our
experiments, we show that the model gains performance by using an increasing
receptive field within the \fusion.

The \sequenceDecoder is the second part of the \sequentialModule that
upsamples the produced \sequenceCode to the desired output resolution
resulting in the \sequenceDecoderOutput. Analogous to the \visualDecoder, we
recurrently perform the upsampling followed by a bottleneck layer and a dense
block to generate a pixel-wise \sequenceDecoderOutput map. To achieve an  
independent data processing of the \sequenceDecoder and
\visualDecoder, we neither share weights between both pathways nor connect
them via skip connections with the encoder of the FCN.

The last building block of the \sequentialModule is the merge layer. Its main
objectives are to merge the visual features with the \sequenceDecoderOutput
and to compute the label mask as the output of the system. First, we
concatenate the input feature volumes along their feature axis and pass the
result to a bottleneck layer using $12$ kernels, where the actual merge takes
place. Then we pass the resulting feature volume through a stack of two 2D
convolutional layers. Finally, we convolve the feature volume into the label
mask using a bottleneck layer with $3$ kernels for respective class labels and
perform a pixel-wise softmax along the feature axis. Further details on the
specific number of layers and parameters can be found in
\figref{fig:archconcept} and \secref{sec:exp}, where we also evaluate the
influence of our key architectural design decisions.


\section{Experimental Evaluation}
\label{sec:exp}
Our experiments are designed to show the capabilities of our method and to
support our key claims, which are: (i)~Our approach generalizes well and
robustly identifies the crops and weeds after the visual appearance of the
plants and the soil has changed without the need for adapting the model
through retraining and (ii)~is able to extract features about the spatial
arrangement of the plantation and is able to perform the \cropvsweed
classification task solely using this geometric information source.

\subsection{Experimental Setup}

\begin{table}[t]
\centering
\vspace{0.15cm}
\caption{Dataset Information}
\begin{tabular}{rC{1.5cm}C{1.5cm}C{1.5cm}}
    \toprule
   & \textbf{Bonn2016} & \textbf{Bonn2017} & \textbf{Stuttgart} \\ \midrule 
   \#~images & 10,036 &  864 & 2,584 \\ 
   crop pixels & 1.7\% & 0.3\% & 1.5\% \\
   approx. crop size & 0.5-100\,cm$^2$ & 0.5-3\,cm$^2$ & 3-25\,cm$^2$ \\
   weed pixels & 0.7\% & 0.05\% & 0.7\% \\
   approx. weed size & 0.5-60\,cm$^2$ & 0.5-3\,cm$^2$ & 2-60\,cm$^2$ \\
 \bottomrule
  \end{tabular}
\label{tab:datasets}

\end{table}

\begin{figure}[t]
\setlength\tabcolsep{4pt}
  \begin{tabular}{ccc}
      \small{Bonn2016} & \small{Bonn2017} & \small{Stuttgart} \\
      {\includegraphics[width=0.30\linewidth]{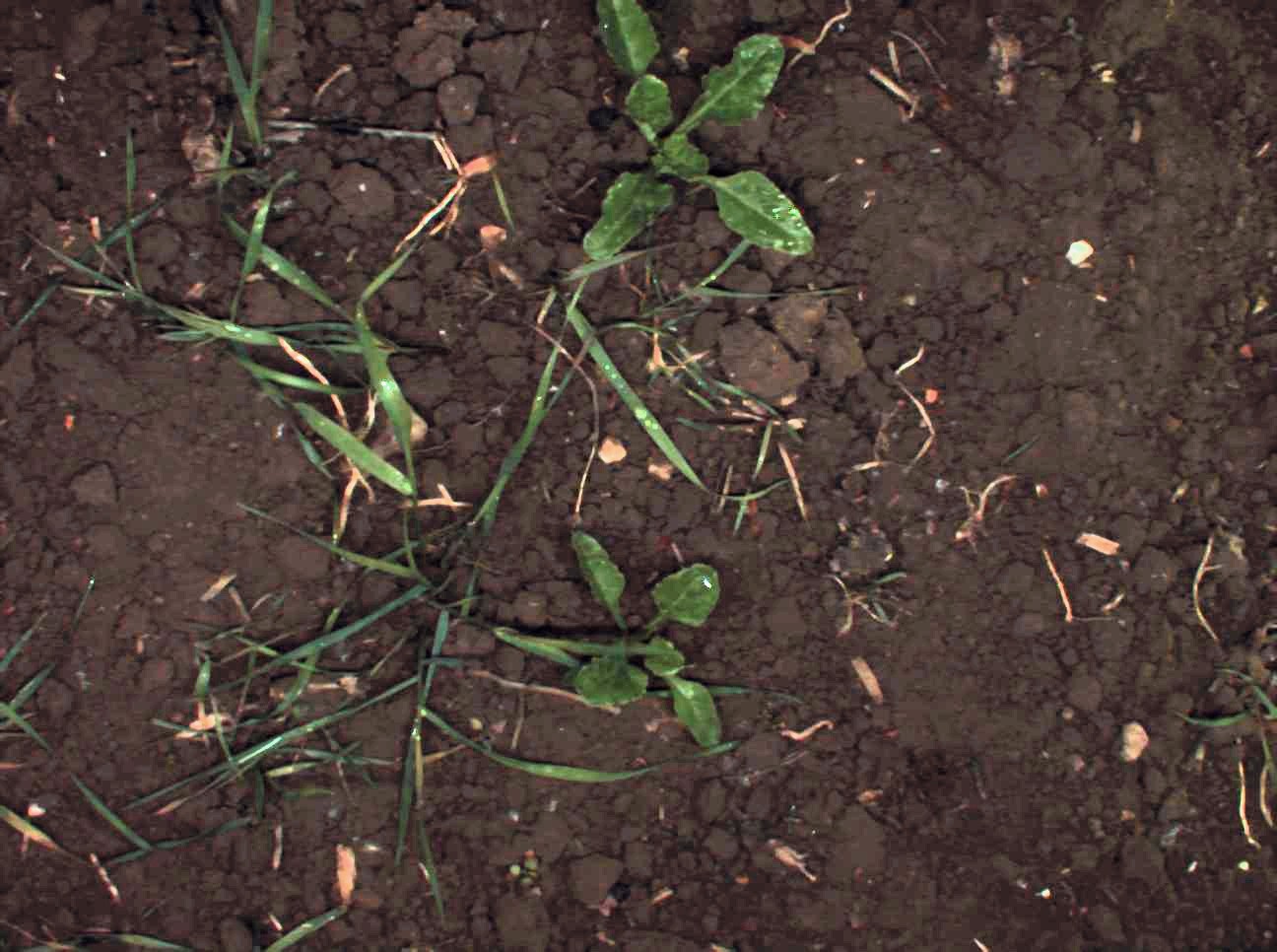}}
   &
      {\includegraphics[width=0.30\linewidth]{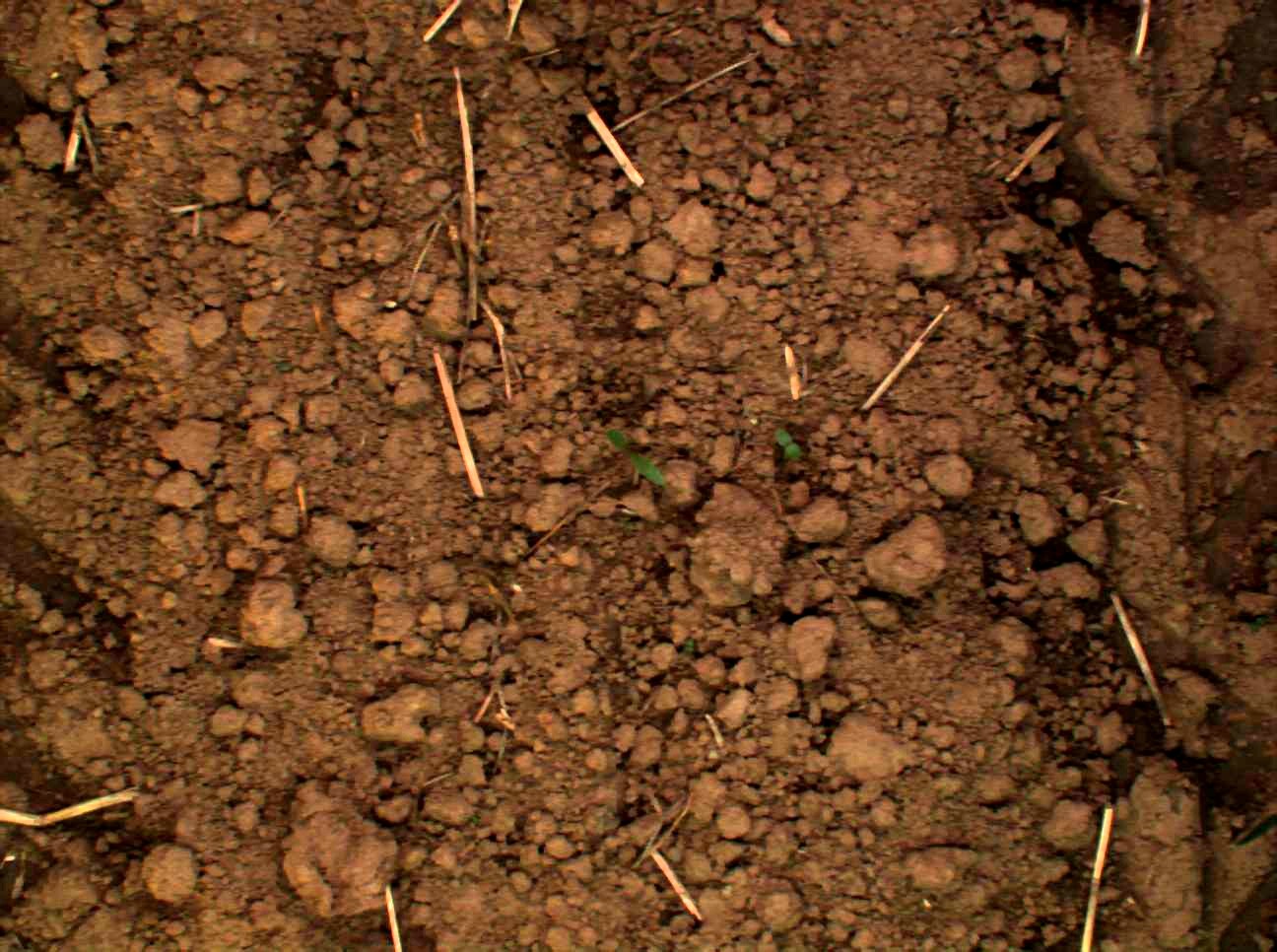}}
   &
      {\includegraphics[width=0.30\linewidth]{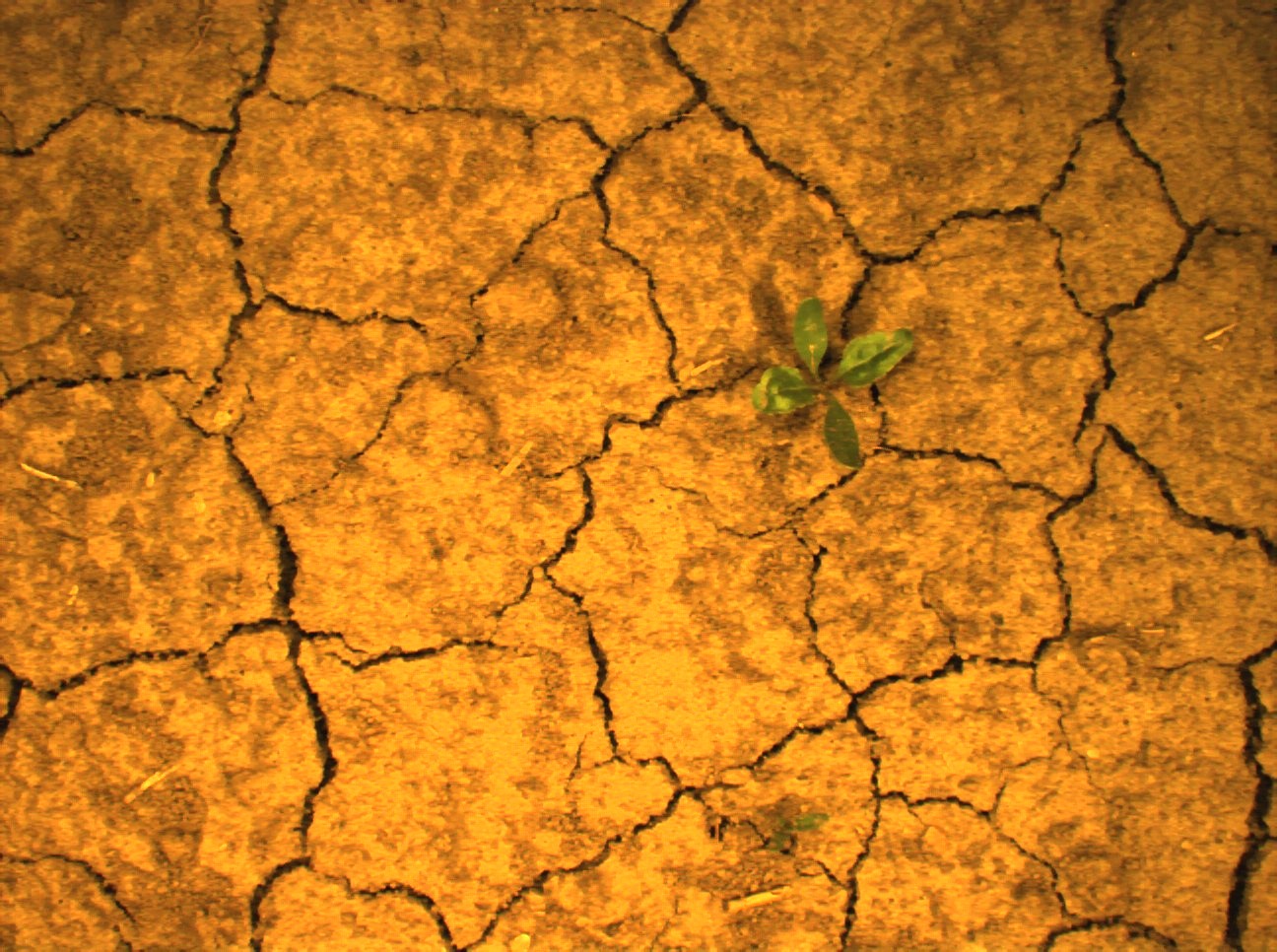}}\\

      {\includegraphics[width=0.30\linewidth]{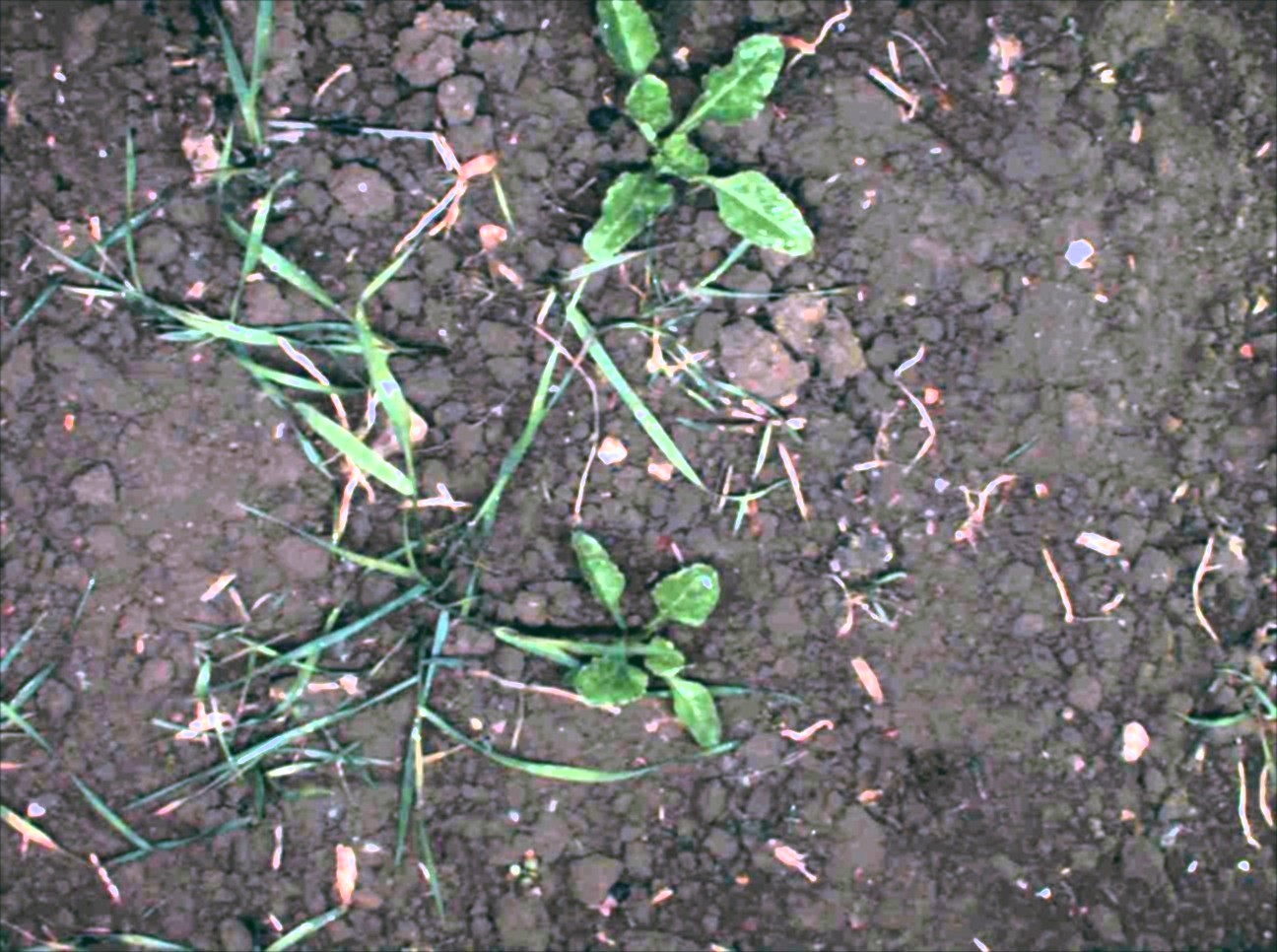}}
   &
      {\includegraphics[width=0.30\linewidth]{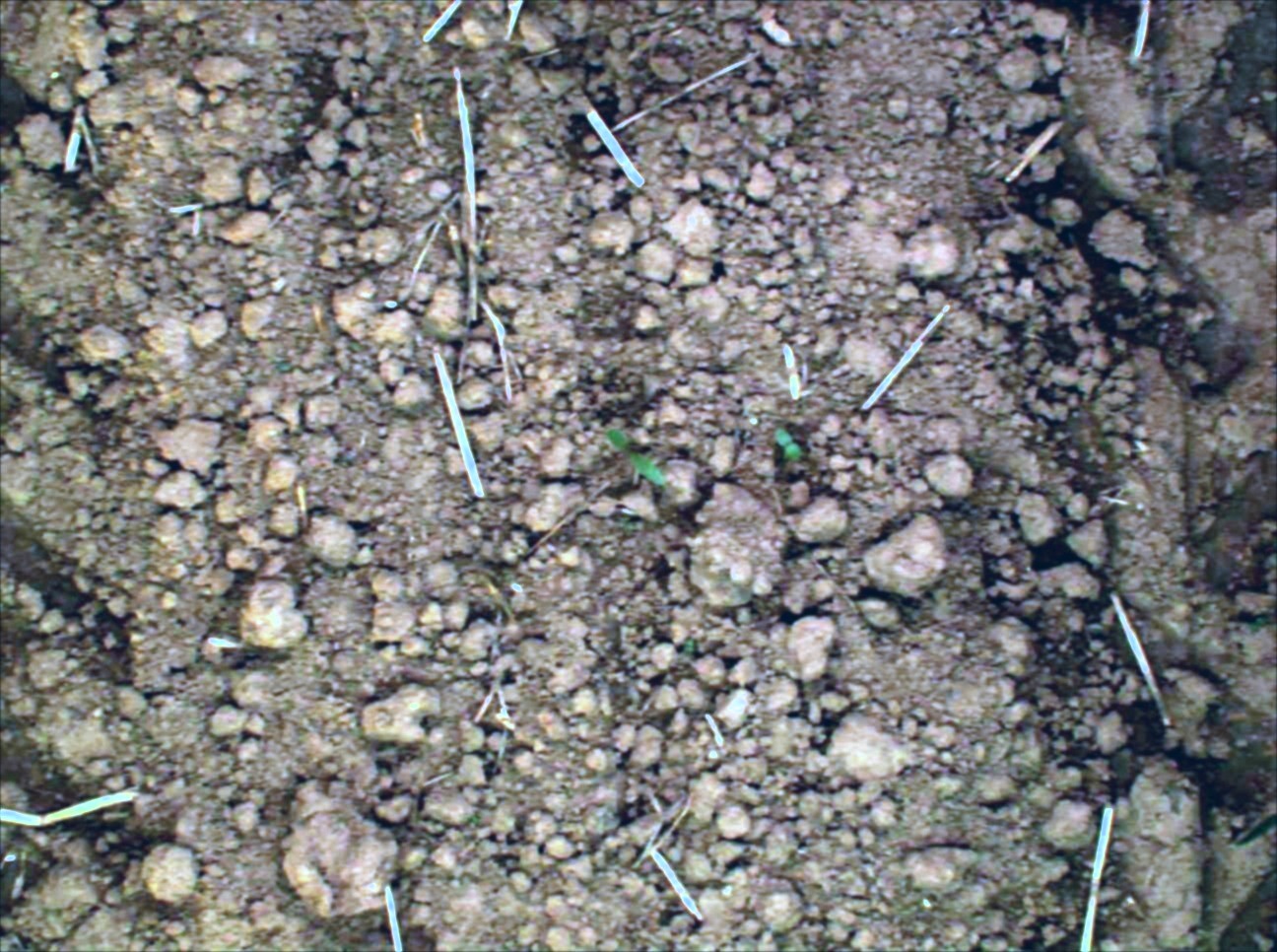}}
   &
      {\includegraphics[width=0.30\linewidth]{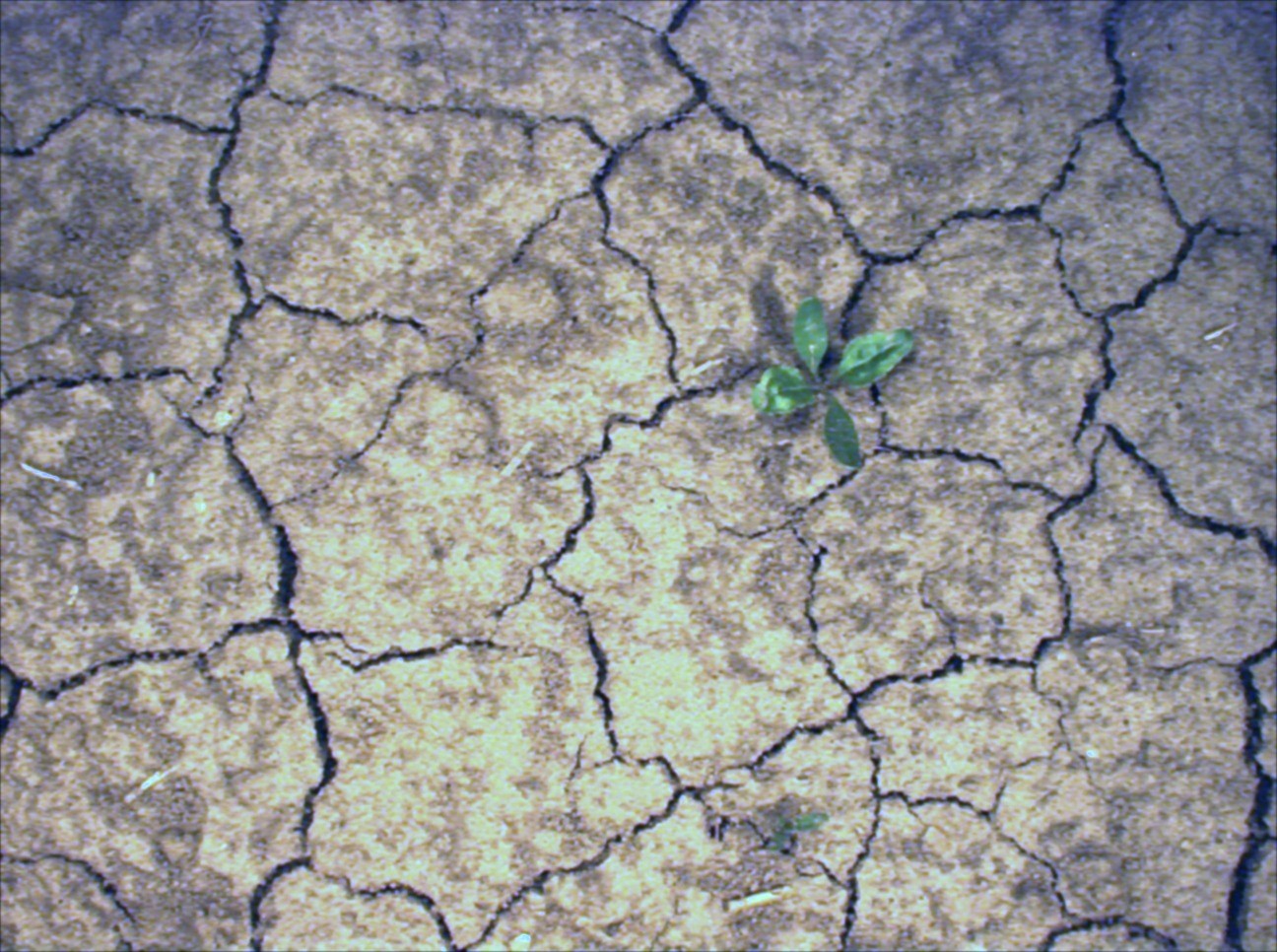}}

  \end{tabular}
\caption{Top: RGB images from each dataset containing different crop growth
stages, weed types, illumination conditions and soil types. Bottom: Example
RGB images after preprocessing.}
\label{fig:dataset}
\vspace{-0.15cm}
\end{figure}

The experiments are conducted on different sugar beet fields located near Bonn
and near Stuttgart in Germany. Each dataset has been recorded with different
variants of the BOSCH DeepField Robotics BoniRob platform. All robots use the
4-channel RGB+NIR camera \mbox{JAI AD-130 GE} mounted in nadir view, but with
different lighting setups. Every dataset contains crops of different growth
stages, while the datasets differ from each other in terms of the weed and
soil types. The dataset collected in Bonn is publicly available and contains
data collected over two month on a sugar beet field. A detailed description of
the data, the setup for its acquisition, and the robot is provided by Chebrolu
\etal~\cite{chebrolu2017ijrr}. 

We evaluate our system in terms of an {object-wise} performance, i.e. a metric
for measuring the performance on plant-level. Here, we compare the predicted
label mask with the crop and weed objects given by the class-wise connected
components from the corresponding ground truth segments. We consider plant
segments with a minimum size of $0.5$\,cm$^2$ in object space and consider
smaller objects to be noise as they are only represented by around 50 pixels
and therefore would bias the performance. In all experiments, we only consider
non-overlapping images to have separate training, validation, and test data.

We refer to the \emph{baseline} approach as our encoder-decoder FCN using the
proposed preprocessing but without the sequential module. We refer to
\emph{our} approach as the combination of the baseline with our proposed
\sequentialModule. Note that we use a comparable number of parameters for the
baseline approach to exclude that the performance is affected by less capacity
of the model.

For comparison, we also evaluate the performance on our recently published
approaches: (i)~using a semi-supervised vision and geometry based approach
based on random forests \cite{lottes2017iros}  and (ii)~employing a purely
visual FCN classifier solely based on RGB data \cite{milioto2018icra}, which
additionally takes vegetation indices as additional plant features (PF) into
account. We refer to these approaches with RF and FCN+PF respectively.

\subsection{Parameters}
\label{sec:implementation_details}

In our experiments, we train all networks from scratch using downsampled
images with a resolution $W=512$ and $H=384$ which yields a ground resolution
of around $1\,\frac{\text{mm}}{\text{pixel}}$. We use a sequence length $S=5$
and use a grow rate $G=4$ for the dense blocks. For training our architecture,
we follow the common best practices: we initialize the weights as proposed by
He \etal~\cite{he2015iccv} and use the RMSPROP optimizer with a mini-batch
size of $B=2$ leading to $10$ images per mini-batch. We use a weighted cross-
entropy loss, where we penalize prediction errors for the crops and weeds by a
factor of $10$. We set the initial learning rate to $0.01$ and divide it by
$10$ after $10$ and $25$ epochs respectively. We stop the training
after $200$ epochs corresponding to a total duration of approx. $58$ hours. We use
dropout with a rate of $\frac{1}{3}$. We implemented our approach using Tensorflow
and it provides classification results with a processing rate of approximately
$5\,$Hz ($200$\,ms) on a {NVIDIA Geforce GTX 1080 Ti} GPU.
We selected the values for $S$, $G$, $B$ and use RMSPROP with the
proposed learning rate schedule as this combination performs best within our
hyperparameter search on the Bonn2016 validation dataset. We choose $S=5$ as
it is a good trade-off between the present spatial information and the
computational cost per sequence.

\subsection{Performance Under Changing Environmental Conditions}

\small
\tabcolsep=0.11cm
\begin{table}[t]
\vspace{0.15cm}
\centering
\caption{Cross-dataset performance (object-wise) when training on Bonn2016 and testing on Stuttgart}
\begin{tabular}{C{2cm}|C{1.5cm}cccc}
    \toprule    
      Approach &   avg. F1\,[\%]  &  \multicolumn{2}{C{2cm}}{Recall\,[\%]} &   \multicolumn{2}{C{2cm}}{Precision\,[\%]}\\ 
               & &        Crop &      Weed           &               Crop &  Weed \\
     \midrule

      our &  \textbf{92.4} & \textbf{95.4} & \textbf{87.8} & \textbf{89.1} & \textbf{97.9}\\
      baseline & 81.9 & 75.2 & 80.0 & 87.7 & 85.9\\
      RF~\cite{lottes2017iros} &  48.0 & 36.4 & 66.5 & 34.7 & 55.4\\
      FCN+PF~\cite{milioto2018icra} & 74.2 & 85.1 & 65.2 & 64.6 & 87.9\\
      \midrule
      RF$^\star$~\cite{lottes2017iros} & 91.8 & 95.1 & 92.2 & 85.2 & 95.4 \\
      FCN+PF$^\star$~\cite{milioto2018icra} & 90.5 & 91.4 & 89.1 & 86.6 & 95.5 \\
      \bottomrule
      \multicolumn{6}{p{0.9\linewidth}}{\scriptsize{$\star$ Approach with \textbf{additional labeling} effort and retraining of classifier.}}
\end{tabular}

\label{tab:generalizationStuttgart}
\vspace{-0.15cm}
\end{table}
\normalsize

The first experiment is designed to support the claim of superior
generalization capabilities with a high performance of our approach, even when
the training and test data notably differ concerning the visual appearance in
the image data. Note that we do not perform retraining in this experiment as
its main purpose is to represent practical challenges for a \cropvsweed
classification, as argued by Slaughter \etal~\cite{slaughter2008cea}. We
evaluate the performance in case that we operate with a different robotic
system in previously unseen fields with changes in the visual appearance
in the image data induced by different weed types and weed pressure and soil
conditions. Therefore, we train the classification model on the entire data of
the {Bonn2016} dataset and evaluate the performance on the {Stuttgart} and
{Bonn2017} datasets. For a fair comparison with RF~\cite{lottes2017iros}, we
do not employ the online learning and provide no labeled data from the
targeted field to initialize the random forest classifier.

\small
\tabcolsep=0.11cm
\begin{table}[t]
\vspace{0.15cm}
\centering
\caption{Cross-dataset performance (object-wise) when training on Bonn2016 and testing on Bonn2017}

\begin{tabular}{C{2cm}|C{1.5cm}cccc}
\toprule
Approach & avg. F1\,[\%]  & \multicolumn{2}{C{2cm}}{Recall\,[\%]} & \multicolumn{2}{C{2cm}}{Precision\,[\%]} \\ 
         &                &         Crop & Weed              & Crop & Weed \\
\midrule
our &  \textbf{86.6} & \textbf{91.2} & \textbf{95.3} & \textbf{90.3} & 72.7\\
baseline & 73.7 & 82.0 & 93.9 & 80.6 & 51.1\\
RF~\cite{lottes2017iros} & 50.2 & 30.5 & 55.7 & 42.7 & \textbf{77.5}\\
FCN+PF~\cite{milioto2018icra} & 67.3 & 83.1 & 47.4 & 92.2 & 46.8\\
\midrule
RF$^\star$~\cite{lottes2017iros} & 92.8 & 96.1 & 92.0 & 86.2 & 97.6 \\
FCN+PF$^\star$~\cite{milioto2018icra} & 89.1 & 95.6 & 69.8 & 92.4 & 85.7 \\
      \bottomrule
            \multicolumn{6}{p{0.9\linewidth}}{\scriptsize{$\star$ Approach with \textbf{additional labeling} effort and retraining of classifier.}}
\end{tabular}
\label{tab:generalizationBonn}
\vspace{-0.15cm}
\end{table}
\normalsize

\tabref{tab:generalizationStuttgart} summarizes the obtained performance on
the Stuttgart dataset. Our approach outperforms all other methods. We detect
more than $95\%$ of the crops and around $88\%$ of the weeds and gain around
$10\%$ performance to the second best in terms of the average F1-score. The
results obtained on the {Bonn2017} dataset confirm this observation as we
observe a margin of around $5\%$ to the second best approach with respect to
the average F1-score. \tabref{tab:generalizationBonn} shows the achieved
performance for the Bonn2017 dataset. Here, our approach detects most of the
plants correctly with a recall of $91\%$ of the crops and $95\%$ of the weeds.

\figref{fig:sequence} illustrates the performance our our approach on the
Stuttgart dataset. Aside from the fact that we perform the classification in a
sequence-to-one fashion, we present the predictions over the whole sequence.
This result supports the high recall obtained for the crop class. Furthermore,
it can be seen that the crops and weed pixels are precisely separated from the
soil, which indicates a high performance for the vegetation separation.

The changes in the visual appearance of the image data of the Stuttgart and
Bonn2017 datasets compared to the Bonn2016 training dataset lead to a
substantial decrease of the obtained performance by the pure visual
approaches, i.e., the baseline and FCN+PF~\cite{milioto2018icra}. The
comparison suggests that the additional exploitation of the arrangement
information leads to a better generalization to other field environments. The
performance of the RF~\cite{lottes2017iros} breaks down due to a wrong
initialization of the geometrical classifier induced by wrong prediction of
the visual classifier in the first iteration.
Note that we executed a t-test for our approach and the respective
baselines in \tabref{tab:generalizationStuttgart} and
\tabref{tab:generalizationBonn}. In all experiments, our approach was
significantly better than the baselines under a 99\% confidence level.

The two bottom rows of the tables reflect the obtained performance for the
FCN+PF$^\star$~\cite{milioto2018icra} and RF$^\star$~\cite{lottes2017iros},
when the classification models have access to small amounts of training data
coming from the target field, e.g., to adapt their parameters to the data
distribution through retraining. Therefore, we manually labeled $100$
additional images from the test dataset and used this data to retrain the
FCN+PF$^\star$~\cite{milioto2018icra} and RF$^\star$~\cite{lottes2017iros}
classifier. Our approach obtains a comparable performance terms of the
average F1-score, i.e., above $90$\,\%, for the Stuttgart dataset. For the
Bonn2017 dataset our approach cannot reach this level of performance. The
results indicate that retraining of classification models leads to a more
reliable performance, but with our approach we make a big step towards
bridging this gap.

Thus, this experiment clearly shows the superior generalization capabilities
of our approach and the impact of exploiting sequential data. However, we will
further investigate the influence of the \sequentialModule and other key
design decision in our ablation study in \secref{sec:ablation_study}.

\subsection{Performance Under Changing Growth Stage}

\tabcolsep=0.11cm
\begin{table}[]
\vspace{0.15cm}
\caption{Generalization capabilities: Test performance on data containing
different growth stages of the sugar beets}
\centering
\begin{tabular}{C{2cm}|C{1.5cm}cccc}
\toprule
Approach & avg. F1\,[\%] &   \multicolumn{2}{C{2cm}}{Recall\,[\%]} &  \multicolumn{2}{C{2cm}}{Precision\,[\%]}\\ 
        &                &  Crop &  Weed                      & Crop & Weed \\
      \midrule
       our & \textbf{92.3} & \textbf{96.1} & \textbf{92.4} & \textbf{96.6} & \textbf{81.5}\\
       baseline & 69.3 & 94.1 & 45.5& 71.4& 77.9\\
      \bottomrule
\end{tabular}
\label{tab:growth}
\end{table}

This experiment is designed to evaluate the generalization capabilities of our
approach regarding changes due the growth stage. This scenario is typical in
practical use cases, where the robot enters the same field again after some
time. We evaluate this by choosing training and test subsets from the
{Bonn2016} datasets with a temporal difference of around $10$ days. The
training data is given by around $1,500$ images acquired in early season at
the two-leave growth stage with an average leaf of around $1$\,cm$^2$.
The test data is given by images from another location in the same field
captured later in time, where the sugar beets have an average size of
$6$\,cm$^2$, i.e. about six times larger than the training examples.

\tabref{tab:growth} shows the obtained performance of our approach in
comparison to the baseline model. With an average F1-score of $91\%$, our
approach provides a solid performance under changing growth stages. The
results show that our approach can exploit the repetitive arrangement pattern
of the sugar beets from the sequential data stream, since it better
generalizes to other growth stages compared to the baseline model. 
The performance of the baseline model suffers from the different visual
appearance as it solely relies on visual cues.  The decrease in performance
of the baseline model underlines that exploiting sequential information
compensates for changes in the visual appearance.

\subsection{Ablation Study}
\label{sec:ablation_study}

\tabcolsep=0.11cm
\begin{table}[t]
\vspace{-0.2cm}
\caption{Ablation Study}
\centering
\begin{tabular}{cl|ccc}
    
    \toprule

      & Approach & 
       {Bonn2016} &
       {Bonn2017} &
       {Stuttgart} \\ 

      \midrule
       
        & Vanilla FCN              & {96.7} & {72.5} & {69.3} \\
       + &Preprocessing            & {96.5} & {79.1} & {77.8} \\
       + &Sequential               & \textbf{97.5} & 83.0  &  89.2 \\
       + &Spatial Context (Our)    & 96.9 & \textbf{86.6} & \textbf{92.4} \\
      \bottomrule

      \multicolumn{5}{p{0.9\linewidth}}{\scriptsize{All classifiers are
      trained on the Bonn2016 training data.}}

\end{tabular}
\label{tab:ablation}
\vspace{-0.2cm}
\end{table}

In this experiment, we show the effect of the most central architectural
design choices resulting in the largest gain and improvement of the generalization
capability. Therefore, we evaluate different architectural configurations of
our approach using 75\% of the {Bonn2016} dataset for training and evaluate
the performance on (i)~20\% held-out testset of Bonn2016, (ii)~Stuttgart and
(iii)~Bonn2017.

\tabref{tab:ablation} reports the obtained object-wise test performance with
respect to the average F1-score. Here, we start with a vanilla FCN
corresponding to our Encoder-Decoder FCN described in \secref{sec:baseFCN}.
Then, we add our proposed preprocessing which helps to generalize to
different fields. It minimizes the effect of different lighting conditions
(see \figref{fig:dataset}) at the cost of a neglectable decrease in
performance on the Bonn2016 dataset. We can furthermore improve the
generalization capabilities by using the sequential
module on top. Adding then more spatial context by increasing the receptive
field of the sequential module by using bigger kernels and dilated
convolutions further improves the performance. 

The high performance of all configurations on the held-out Bonn2016 data
indicates that FCNs generally obtain a stable and high performance under a
comparably low diversity in the data distribution. Thus, diversity between the
training and test data is crucial to evaluate \cropvsweed
classification system under practical circumstances. We conclude that
preprocessing and exploitation of the repetitive pattern given by the crop
arrangement helps to improve the generalization capabilities of our sequential
FCN-based
\cropvsweed classification system.

\subsection{Experiment on Learning the Crop-Weed Arrangement}
\label{sec:simulation}

\begin{figure}
 \vspace{0.15cm}
 \centering
 
 \includegraphics[width=1.0\linewidth]{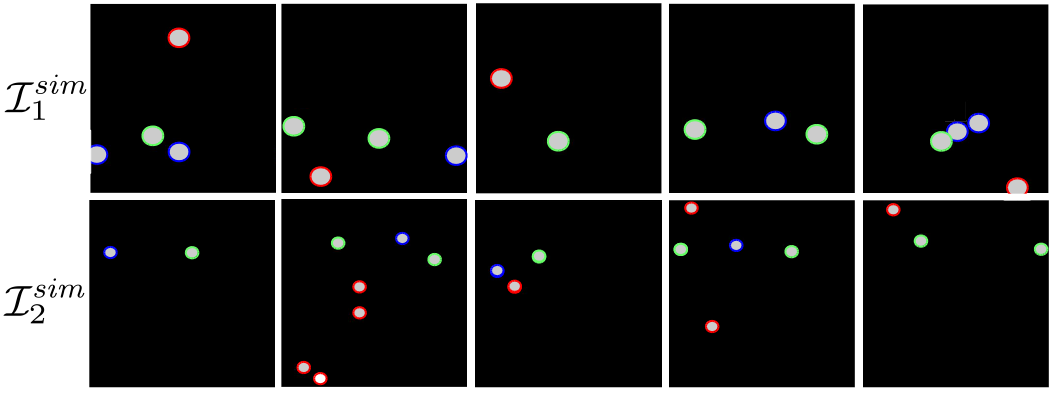}

 \caption{Two realizations of simulated image sequences
 $\mathcal{I}^{\mathit{sim}}_1$ and $\mathcal{I}^{\mathit{sim}}_2$ with
 different properties as mentioned in \secref{sec:simulation}. The blob size
 is uniform within a sequence. The colored contours refer to the corresponding
 ground truth labels, i.e. crop (green), weed (red) and  intra-row weed
 (blue).}
  \label{fig:simulateddata}
  \vspace{-0.05cm}
\end{figure}

Using experiments with simulated data, we finally demonstrate the capability
of our approach to extract information about the spatial arrangement of the
crops and weeds. For this purpose, we create input data providing
\emph{solely} the signal of the plant arrangement as a potential information
to distinguish the crop and weeds.

We render binary image sequences encoding the crops and weeds as uniform sized
blobs while imitating (i)~the crop-weed arrangement and (ii)~the acquisition
setup of the robotic system in terms of the camera setup and its motion in
space. \figref{fig:simulateddata} depicts two simulated sequences and their
corresponding ground truth information. Note that within a sequence, we only
model blobs of uniform size. Thus, we provide neither spectral nor shape
information as a potential input signal for the classification task. We
hypothesize that an algorithm solving the classification task given the
simulated data is also able to learn features describing the arrangement of
crops and weeds.

We furthermore explicitly simulate intra-row weeds, i.e., weeds located close
to the crop row. To properly detect intra-row weeds, the classifier needs to
learn the intra-row distance between crops instead of solely memorizing that
all vegetation growing close to the center line of a crop row belongs to the
crop class. We model the following properties for creating the simulated
sequences, as shown in \figref{fig:simulateddata}: (1) intra-row distance
($15$-$25$\,cm), (2) inter-row distance ($30$-$60$\,cm), (3) weed pressure
($0$-$200$\% w.r.t. the crops present in the sequence), (4) plant size
($0.5$-$8.0$\,cm$^2$), and (5) the camera motion along the row by a slight
variation of the steering angle. We pertubate the plant
locations by Gaussian noise ($\mu = 0$ and $\sigma^2 = 10\%$ of the intra-row distance).

\small
\tabcolsep=0.11cm
\begin{table}[]
\vspace{0.15cm}
\caption{Test performance on simulated data}
\centering
\begin{tabular}{c|cccc|ccc}
    \toprule

      \multirow{2}{*}{Approach} & 
      avg. & 
      \multicolumn{3}{c}{Recall\,[\%]} &
      \multicolumn{3}{c}{Precision\,[\%]}\\ 

        &
       F1\,[\%] &
      Crop & 
      Weed & 
      Intra-Weed & 
      Crop & 
      Weed & 
      Intra-Weed \\

  \midrule

      our& \textbf{93.6} &\textbf{95.4}&\textbf{99.1}&\textbf{84.2}&\textbf{98.3}&\textbf{99.1}&\textbf{85.5}\\
      baseline& 47.3&  61.1&59.3&9.4&75.9&80.4&4.4\\

      \bottomrule
\end{tabular}
\label{tab:simulation}
\end{table}
\normalsize

\tabref{tab:simulation} shows the performance of our  approach
compared to the baseline approach. These results demonstrate that our approach
can exploit the sequential information to extract the pattern of the crop
arrangement for the classification task. In contrast, the baseline model is
unable to properly identify the crops and weeds due to missing shape
information.

\section{Conclusion}
\label{sec:conclusion}

In this paper, we presented a novel approach for precision agriculture robots
that  provides a pixel-wise semantic segmentation into crop and weed. It
exploits a fully convolutional network integrating sequential information. We
proposed to encode the spatial arrangement of plants in a row using 3D
convolutions over an image sequence. Our thorough experimental evaluation
using real-world data demonstrates that our system (i)~generalizes better to
unseen fields in comparison to other state-of-the-art approaches and~(ii) is
able to robustly classify crop in different growth stages. Finally, we show
that the proposed \sequentialModule actually encodes the spatial arrangement
of the plants through simulation.

\bibliographystyle{IEEEtran}
\bibliography{glorified}

\begin{thebibliography}{10}
\providecommand{\url}[1]{#1}
\csname url@rmstyle\endcsname
\providecommand{\newblock}{\relax}
\providecommand{\bibinfo}[2]{#2}
\providecommand\BIBentrySTDinterwordspacing{\spaceskip=0pt\relax}
\providecommand\BIBentryALTinterwordstretchfactor{4}
\providecommand\BIBentryALTinterwordspacing{\spaceskip=\fontdimen2\font plus
\BIBentryALTinterwordstretchfactor\fontdimen3\font minus
  \fontdimen4\font\relax}
\providecommand\BIBforeignlanguage[2]{{%
\expandafter\ifx\csname l@#1\endcsname\relax
\typeout{** WARNING: IEEEtran.bst: No hyphenation pattern has been}%
\typeout{** loaded for the language `#1'. Using the pattern for}%
\typeout{** the default language instead.}%
\else
\language=\csname l@#1\endcsname
\fi
#2}}

\bibitem{mueter2013aec}
M.~M{\"u}ter, P.~S. Lammers, and L.~Damerow, ``Development of an intra-row
  weeding system using electric servo drives and machine vision for plant
  detection,'' in \emph{Proc.~of the International Conf.~on Agricultural
  Engineering LAND.TECHNIK (AgEng)}, 2013.

\bibitem{Langsenkamp2014cigr}
F.~Langsenkamp, F.~Sellmann, M.~Kohlbrecher, A.~Kielhorn, W.~Strothmann,
  A.~Michaels, A.~Ruckelshausen, and D.~Trautz, ``Tube stamp for mechanical
  intra-row individual plant weed control,'' in \emph{Proc.~of the
  International Conf.~of Agricultural Engineering (CIGR)}, 09 2014.

\bibitem{haug2014wacv}
S.~Haug, A.~Michaels, P.~Biber, and J.~Ostermann, ``{Plant Classification
  System for Crop / Weed Discrimination without Segmentation},'' in
  \emph{Proc.~of the IEEE Winter Conf.~on Applications of Computer Vision
  (WACV)}, 2014.

\bibitem{lottes2017icra}
\BIBentryALTinterwordspacing
P.~Lottes, R.~Khanna, J.~Pfeifer, R.~Siegwart, and C.~Stachniss, ``{UAV-Based
  Crop and Weed Classification for Smart Farming},'' in \emph{Proc.~of the IEEE
  Intl.~Conf.~on Robotics \& Automation (ICRA)}, 2017.
\BIBentrySTDinterwordspacing

\bibitem{milioto2018icra}
\BIBentryALTinterwordspacing
A.~Milioto, P.~Lottes, and C.~Stachniss, ``{Real-time Semantic Segmentation of
  Crop and Weed for Precision Agriculture Robots Leveraging Background
  Knowledge in CNNs},'' in \emph{Proc.~of the IEEE Intl.~Conf.~on Robotics \&
  Automation (ICRA)}, 2018.
\BIBentrySTDinterwordspacing

\bibitem{mortensen2016cigr}
A.~K. Mortensen, M.~Dyrmann, H.~Karstoft, R.~N. J{\"o}rgensen, and R.~Gislum,
  ``{Semantic Segmentation of Mixed Crops using Deep Convolutional Neural
  Network},'' in \emph{Proc.~of the International Conf.~of Agricultural
  Engineering (CIGR)}, 2016.

\bibitem{nieuwenhuizen2009phd}
A.~Nieuwenhuizen, ``Automated detection and control of volunteer potato
  plants,'' Ph.D. dissertation, Wageningen University, 2009.

\bibitem{potena2016ias}
C.~Potena, D.~Nardi, and A.~Pretto, ``Fast and accurate crop and weed
  identification with summarized train sets for precision agriculture,'' in
  \emph{Proc. of Int.~Conf.~on Intelligent Autonomous Systems (IAS)}, 2016.

\bibitem{slaughter2008cea}
D.~Slaughter, D.~Giles, and D.~Downey, ``Autonomous robotic weed control
  systems: A review,'' \emph{Computers and Electronics in Agriculture},
  vol.~61, no.~1, pp. 63 -- 78, 2008.

\bibitem{chebrolu2017ijrr}
N.~Chebrolu, P.~Lottes, A.~Schaefer, W.~Winterhalter, W.~Burgard, and
  C.~Stachniss, ``{Agricultural Robot Dataset for Plant Classification,
  Localization and Mapping on Sugar Beet Fields},'' \emph{Intl.~Journal~of
  Robotics Research (IJRR)}, 2017.

\bibitem{lottes2016jfr}
P.~Lottes, M.~Hoeferlin, S.~Sanders, and C.~Stachniss, ``{Effective
  Vision-Based Classification for Separating Sugar Beets and Weeds for
  Precision Farming},'' \emph{Journal of Field Robotics (JFR)}, 2016.

\bibitem{cicco2017iros}
\BIBentryALTinterwordspacing
M.~Cicco, C.~Potena, G.~Grisetti, and A.~Pretto, ``{Automatic Model Based
  Dataset Generation for Fast and Accurate Crop and Weeds Detection},'' in
  \emph{Proc.~of the IEEE/RSJ Intl.~Conf.~on Intelligent Robots and Systems
  (IROS)}, 2017.
\BIBentrySTDinterwordspacing

\bibitem{mccool2017ral}
C.~McCool, T.~Perez, and B.~Upcroft, ``{Mixtures of Lightweight Deep
  Convolutional Neural Networks: Applied to Agricultural Robotics},''
  \emph{IEEE Robotics and Automation Letters (RA-L)}, 2017.

\bibitem{milioto2017uavg}
\BIBentryALTinterwordspacing
A.~Milioto, P.~Lottes, and C.~Stachniss, ``{Real-time Blob-wise Sugar Beets vs
  Weeds Classification for Monitoring Fields using Convolutional Neural
  Networks},'' in \emph{Proc.~of the Intl.~Conf.~on Unmanned Aerial Vehicles in
  Geomatics}, 2017.
\BIBentrySTDinterwordspacing

\bibitem{szegedy2016cvpr}
\BIBentryALTinterwordspacing
C.~Szegedy, V.~Vanhoucke, S.~Ioffe, J.~Shlens, and Z.~Wojna, ``{Rethinking the
  Inception Architecture for Computer Vision},'' in \emph{Proc.~of the IEEE
  Conf.~on Computer Vision and Pattern Recognition (CVPR)}, 2016.
\BIBentrySTDinterwordspacing

\bibitem{long2015cvpr-fcnf}
\BIBentryALTinterwordspacing
J.~Long, E.~Shelhamer, and T.~Darrell, ``{Fully Convolutional Networks for
  Semantic Segmentation},'' in \emph{Proc.~of the IEEE Conf.~on Computer Vision
  and Pattern Recognition (CVPR)}, 2015.
\BIBentrySTDinterwordspacing

\bibitem{badrinarayanan2017pami}
V.~Badrinarayanan, A.~Kendall, and R.~Cipolla, ``{SegNet: A Deep Convolutional
  Encoder-Decoder Architecture for Image Segmentation},'' \emph{IEEE Trans.~on
  Pattern Analalysis and Machine Intelligence (TPAMI)}, vol.~39, no.~12, pp.
  2481--2495, 2017.

\bibitem{paszke2016arxiv}
\BIBentryALTinterwordspacing
A.~Paszke, A.~Chaurasia, S.~Kim, and E.~Culurciello, ``{ENet: Deep Neural
  Network Architecture for Real-Time Semantic Segmentation},'' \emph{arXiv
  preprint}, vol. abs/1606.02147, 2016.
\BIBentrySTDinterwordspacing

\bibitem{ronneberger2015micc}
\BIBentryALTinterwordspacing
O.~Ronneberger, P.Fischer, and T.~Brox, ``{U-Net: Convolutional Networks for
  Biomedical Image Segmentation},'' in \emph{Medical Image Computing and
  Computer-Assisted Intervention}, ser. LNCS, vol. 9351.\hskip 1em plus 0.5em
  minus 0.4em\relax Springer, 2015, pp. 234--241.
\BIBentrySTDinterwordspacing

\bibitem{sa2018ral}
\BIBentryALTinterwordspacing
I.~Sa, Z.~Chen, M.~Popvic, R.~Khanna, F.~Liebisch, J.~Nieto, and R.~Siegwart,
  ``{weedNet: Dense Semantic Weed Classification Using Multispectral Images and
  MAV for Smart Farming},'' \emph{IEEE Robotics and Automation Letters (RA-L)},
  vol.~3, no.~1, pp. 588--595, 2018.
\BIBentrySTDinterwordspacing

\bibitem{wendel2016icra}
A.~Wendel and J.~Underwood, ``{Self-Supervised Weed Detection in Vegetable
  Crops Using Ground Based Hyperspectral Imaging},'' in \emph{Proc.~of the IEEE
  Intl.~Conf.~on Robotics \& Automation (ICRA)}, 2016.

\bibitem{lottes2017iros}
\BIBentryALTinterwordspacing
P.~Lottes and C.~Stachniss, ``Semi-supervised online visual crop and weed
  classification in precision farming exploiting plant arrangement,'' in
  \emph{Proc.~of the IEEE/RSJ Intl.~Conf.~on Intelligent Robots and Systems
  (IROS)}, 2017.
\BIBentrySTDinterwordspacing

\bibitem{hall2017iros}
\BIBentryALTinterwordspacing
D.~Hall, F.~Dayoub, T.~Perez, and C.~McCool, ``{A Transplantable System for
  Weed Classification by Agricultural Robotics},'' in \emph{Proc.~of the
  IEEE/RSJ Intl.~Conf.~on Intelligent Robots and Systems (IROS)}, 2017.
\BIBentrySTDinterwordspacing

\bibitem{hall2017icra}
\BIBentryALTinterwordspacing
D.~Hall, F.~Dayoub, J.~Kulk, and C.~McCool, ``{Towards Unsupervised Weed
  Scouting for Agricultural Robotics},'' in \emph{Proc.~of the IEEE
  Intl.~Conf.~on Robotics \& Automation (ICRA)}, 2017.
\BIBentrySTDinterwordspacing

\bibitem{huang2017cvpr-dccn}
\BIBentryALTinterwordspacing
G.~Huang, Z.~Liu, L.~v.~d. Maaten, and K.~Q. Weinberger, ``{Densely Connected
  Convolutional Networks},'' in \emph{Proc.~of the IEEE Conf.~on Computer
  Vision and Pattern Recognition (CVPR)}, 2017.
\BIBentrySTDinterwordspacing

\bibitem{jegou2017arxiv}
\BIBentryALTinterwordspacing
S.~J{\'{e}}gou, M.~Drozdzal, D.~V{\'{a}}zquez, A.~Romero, and Y.~Bengio, ``{The
  One Hundred Layers Tiramisu: Fully Convolutional DenseNets for Semantic
  Segmentation},'' \emph{arXiv preprint:1611.09326}, 2017.
\BIBentrySTDinterwordspacing

\bibitem{srivastava2014jmlr}
\BIBentryALTinterwordspacing
N.~Srivastava, G.~Hinton, A.~Krizhevsky, I.~Sutskever, and R.~Salakhutdinov,
  ``{Dropout: A Simple Way to Prevent Neural Networks from Overfitting},''
  \emph{Journal of Machine Learning Research (JMLR)}, vol.~15, pp. 1929--1958,
  2014.
\BIBentrySTDinterwordspacing

\bibitem{dumoulin2016arxiv}
\BIBentryALTinterwordspacing
V.~Dumoulin and F.~Visin, ``{A guide to convolution arithmetic for deep
  learning},'' \emph{arXiv preprint}, 2018.
\BIBentrySTDinterwordspacing

\bibitem{he2015iccv}
\BIBentryALTinterwordspacing
K.~He, X.~Zhang, S.~Ren, and J.~Sun, ``Delving deep into rectifiers: Surpassing
  human-level performance on imagenet classification,'' in \emph{Proc.~of the
  IEEE Intl.~Conf.~on Computer Vision (ICCV)}, 2015.
\BIBentrySTDinterwordspacing

\end{thebibliography}

\end{document}